# Learning to segment with limited annotations: Self-supervised pretraining with regression and contrastive loss in MRI

Lavanya Umapathy[1,2], Zhiyang Fu[1,2], Rohit Philip[2], Diego Martin[3], Maria Altbach[2], and Ali Bilgin[1,2,4]

[1]Dept. of Electrical and Computer Engineering, University of Arizona, Tucson, USA
[2]Dept. of Medical Imaging, University of Arizona, Tucson, USA
[3]Dept. of Radiology, Houston Methodist Hospital, Houston, USA
[4]Dept. of Biomedical Engineering, University of Arizona, Tucson, USA

*Abstract*—Obtaining manual annotations for large datasets for supervised training of deep learning (DL) models is challenging. The availability of large unlabeled datasets compared to labeled ones motivate the use of self-supervised pretraining to initialize DL models for subsequent segmentation tasks. In this work, we consider two pre-training approaches for driving a DL model to learn different representations using: a) regression loss that exploits spatial dependencies within an image and b) contrastive loss that exploits semantic similarity between pairs of images. The effect of pretraining techniques is evaluated in two downstream segmentation applications using Magnetic Resonance (MR) images: a) liver segmentation in abdominal T2-weighted MR images and b) prostate segmentation in T2-weighted MR images of the prostate. We observed that DL models pretrained using self-supervision can be finetuned for comparable performance with fewer labeled datasets. Additionally, we also observed that initializing the DL model using contrastive loss based pretraining performed better than the regression loss.

## I. INTRODUCTION

Supervised deep-learning (DL) models, trained on large, labeled datasets, have shown state-of-the-art performance in several medical image segmentation tasks [1]. However, obtaining manual annotations for supervised training of DL models can be labor-intensive, requiring countless expert hours. Although obtaining large, annotated datasets can be challenging in MR imaging applications, the availability of large unlabeled datasets motivates the use of self-supervised learning (SSL) techniques. SSL based pretraining has shown promising results in image segmentation tasks [2,3] without the need for extensive labeling. Driven by the choice of an appropriate loss function, SSL can learn representations that can provide a suitable initialization for the data-hungry DL models. These learned representations may be used subsequently for a variety of tasks including image segmentation with few labeled data.

In this work, we explore two different SSL approaches for driving a DL model to learn different feature representations: a) a regression loss (using an image reconstruction task) to exploit spatial dependencies within images and b) a contrastive loss to exploit semantic similarity between pairs of images. A DL model is pretrained using the two loss functions. The effect of the pretraining techniques is evaluated in two different MR imaging applications: a) liver segmentation in T2-weighted radial fast spin echo images of the abdomen b) prostate segmentation in T2-weighted images of the prostate. We perform experiments to evaluate the performance of the self-supervised pretraining techniques under limited labeled data conditions in comparison to a supervised DL baseline model trained on all available labeled data.

## II. METHODS

*A. Deep Learning Models*

Consider a DL model $\psi_\theta(.)$, parametrized by $\theta$, that generates representations $\psi_\theta(x)$ for an input image $x$. During the pretraining stage, we aim to learn the model weights $\theta$ such that the representations generated by the DL model minimize (or maximize) the loss function of choice. The space in which the loss function is computed is separated from this representation space using a projection head $g(.)$, similar to [6]. This projection head transforms the representations $\psi_\theta(.)$ into $\Psi_\theta(.)$ such that

$$\Psi_\theta(x) = g(\psi_\theta(x))$$

During the subsequent finetuning stage for a specific downstream segmentation task, the projection head is discarded, and the DL model is optimized for the segmentation loss function of choice.

*B. Regression Loss*

The regression-based pretraining of the DL model uses an image reconstruction loss to generate representations that exploit the spatial dependencies between pixels. This is in some ways similar to some self-supervised denoising auto-encoder methods (e.g., Noise2Void [14]). Figure 1 shows an illustration of the regression loss based pretraining. For a reference image $x$ let $M$ be a mask of random pixels (a percentage of the input pixels in the image) that are to be replaced by Gaussian noise $N(0, \sigma = 0.01)$. The corrupted image is given by

$$\hat{x} = ((1 - M) \odot x) + N \odot M$$

Here, the operator ⊙ defines an element-wise multiplication operation. The objective is to minimize the L1 reconstruction loss between the reference image and the reconstructed/ denoised image yielded by the representation $\Psi(\hat{x})$.

$$l_\theta(x) = \left\|(M \odot \Psi_\theta(\hat{x})) - (M \odot x)\right\|_1$$

This regression loss is only evaluated at corrupted pixels to prevent the CNN from learning identity relationship. The hypothesis behind the loss function is that to be able to successfully recover a random corrupted pixel, a DL model would need an understanding of the spatial context of the pixel.

*C. Contrastive Loss*

A form of representational learning, contrastive loss learns representations/features that approximate the semantic similarity between pairs of images. Contrastive loss aims to push the representations of images/regions with similar semantic content closer to each other while pushing away the representations of dissimilar images [4,5]. This is illustrated in Figure 2.

For an anchor image $x_a$, its positively similar image $x_p$ and other negative images $x_k^-$, contrastive loss trains the DL model to generate representations $\Psi_\theta(x_a)$ that are closer to $\Psi_\theta(x_p)$ while being away from $\Psi_\theta(x_k^-)$. Similar images could be an image with two different data augmentation schemes applied or two images with similar anatomical content. Several works use a cosine similarity-based distance metric in this representational space [7-8].

The contrastive loss used in this work is defined as follows:

$$l_\theta(x_a, x_p) = -log\left(\frac{e^{D(\Psi_\theta(x_a),\Psi_\theta(x_p))/\tau}}{\sum_{k=1}^{N} e^{D(\Psi_\theta(x_a),\Psi_\theta(x_k))/\tau}}\right)$$

Here, $D(.)$ measures the normalized cosine similarity between two representations, $\tau$ is a temperature scaling parameter, and $N$ is the number of images under consideration. In this work, $\tau = 0.1$ is used. The encoder pathway is pretrained separately to contrast the global content of the images similar to [6]. Similar images are generated using data augmentations schemes (image crop, contrast, and brightness adjustments). The decoder pathway is pretrained to contrast local content of the images similar to [9] (corresponding local patches in the representations of similar images contrasted against random local patches).

*D. Data*

*1) T2-weighted abdominal MR images:* With the approval of the local institutional review board, images were acquired at 1.5T MR imaging scanner (Skyra; Siemens) on 138 subjects using the 2D radial turbo spin echo sequence [15]. The sequence parameters are as follows: field of view=40cm, echo train length=32, echo spacing=6.7ms, base resolution=256, radial views=192, TR=2500ms, slice thickness=8mm, number of slices=21, breath hold time=18s. Composite images were reconstructed using all radial k-space data. The processing pipeline consisted of removal of streak artifacts in the radial acquisition, B1 field correction followed by image normalization to [0,1] scale. De-streaking was performed using either CACTUS [10] or manual removal of streak inducing coils. Of these, 107 subjects were used for pretraining the DL model. A subset of these subjects (N=31) was previously labeled for a liver segmentation study [11] and was split into training (N=21), validation (N=2), and test (n=8) sets for the finetuning stage.

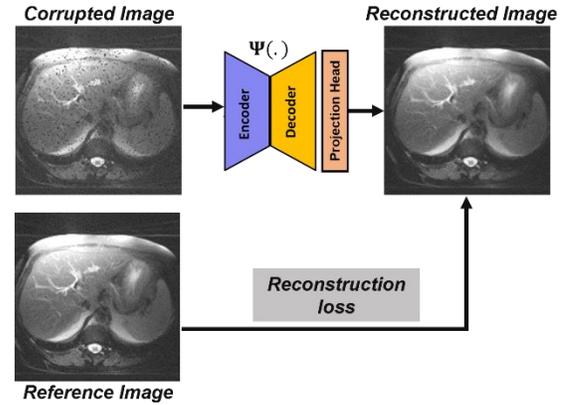

Figure 1: Regression loss based self-supervised pretraining of a deep learning model. A reference image is corrupted by randomly replacing pixels in the image with Gaussian Noise. A deep learning model is trained to minimize the L1 loss between the reference and the reconstructed image. Loss is only evaluated at the corrupted pixel locations.

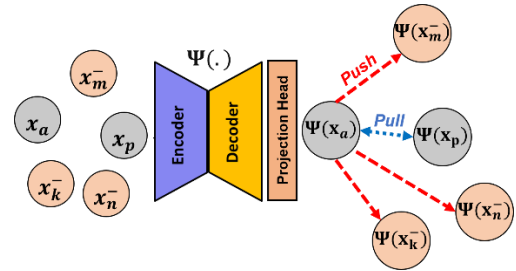

Figure 2: Illustration of contrastive loss for a pair of similar images $x_a$ and $x_p$. Contrastive loss pulls the representations of similar images closer in the representational space while pushing away the representations of dissimilar images.

*2) T2-weighted prostate MR images:* 2D T2-weighted images of the prostate, from the Medical Decathlon Segmentation Challenge for the prostate were used [12] (http://medicaldecathlon.com/). All unlabeled images were used for pretraining the DL model (N=48). For the finetuning stage, a subset of the labeled volumes was split into training (N=15), testing (N=15), and validation sets (N=3).

*E. Experiments*

In this work, a 2D UNET-like [16] architecture was used. The DL model was first pretrained using the two self-supervised learning approaches with unlabeled images corresponding to the two segmentation tasks. For the regression loss based pretraining, 10% of the pixels in the input image were randomly corrupted using Gaussian Noise. Once pretrained, the models were finetuned for the subsequent image segmentation tasks, using a multi-label Dice loss. The number of labeled training images were varied ($N = 1, 2, 4, 8$ etc.) and the performance of the trained DL model was evaluated on the test cohorts using Dice scores. The Dice scores were averaged over multiple runs/ training data combinations. A supervised baseline performance for each segmentation task was also obtained by training the DL model using all available labeled training images for the task. Extensive data augmentation, including random translation, crops, and elastic deformations, was carried out for each finetuning experiment.

All implementation was done in Python using Keras with Tensorflow [13] backend on a Linux system with Titan P100 (NVIDIA) GPUs. For pretraining: epochs=1000, initial learning rate=0.001, optimizer=Adam. For finetuning: epochs=500, initial learning rate=0.001, optimizer=Adam. In all experiments, the learning rate was halved whenever the validation loss plateaued.

## III. RESULTS

### A. Liver Segmentation

Figure 3A compares the segmentation performances (mean Dice scores on the test cohort) of the regression- and the contrastive-loss based pretraining on the T2-weighted abdominal images for the liver segmentation task as the number of training subjects are varied. We observe that the proposed self-supervised pretraining methods approach the performance of the supervised benchmark with very few labeled datasets.

With 8 annotated volumes (30% of the labeled data), the contrastive loss based pretraining achieved mean Dice scores (N=0.89) comparable to the fully supervised baseline performance of the DL model (N=0.89). With only 4 annotated volumes, the contrastive loss based pretraining yielded mean Dice scores (N=0.86) that were within 2% of the supervised baseline performance. In addition, we also observed that on the liver segmentation task, the contrastive loss based pretraining performed better than the regression loss based pretraining. Figure 3B compares the average Dice scores on the test cohort for the DL model when the model is pretrained with contrastive loss (blue) vs when the weights of the DL model is randomly initialized (red). In this case, the use of self-supervised pretraining using contrastive loss helped improve the baseline performance of the DL model. A qualitative comparison of the DL model's segmentation performance when trained with 4 and 8 training volumes, respectively, is shown in Figure 4.

### B. Prostate Segmentation

We observe similar trends in the prostate segmentation task (Figure 5 and 6) where a) with fewer annotated volumes (N=8), the DL model pretrained on contrastive loss yielded mean Dice scores within 3% of the supervised baseline performance and b) contrastive loss performed better than regression loss based pretraining.

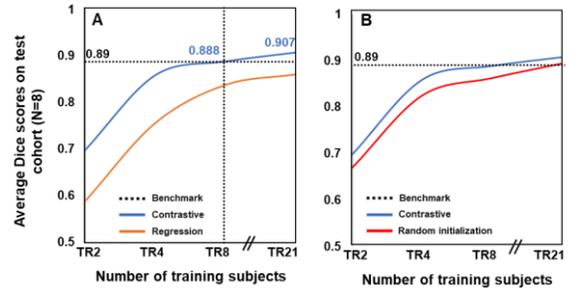

Figure 3: Comparison of segmentation performance (mean Dice scores on the test cohort) of the (A) two self-supervised pretraining techniques on liver segmentation task and (B) contrastive loss based pretrained weights vs random initialization of the deep learning model. The Dice scores are averaged over the test cohort (N=8). The dotted line represents the average Dice score from a supervised baseline deep learning model trained on all available labeled training data (N=21). With fewer labeled datasets, contrastive loss can yield comparable performance to a fully supervised baseline deep learning model.

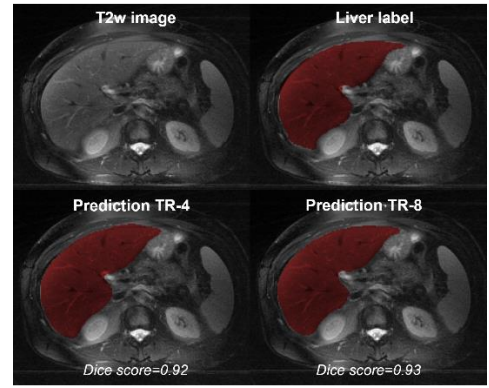

Figure 4: Liver segmentation. Qualitative comparison of liver mask predictions (overlaid in red on T2-weighted abdomen images) from deep learning model pretrained with contrastive loss and finetuned with four (bottom left) and eight (bottom right) training subjects respectively. The Dice scores for the 3D volume for each case is also displayed along each image.

## IV. DISCUSSION

In this work, we explored the potential of self-supervised learning (SSL) pretraining with large unlabeled images using two different loss functions: regression loss and contrastive loss. Each loss function was designed to affect the representations generated by the deep learning (DL) model in different ways. The regression based pretraining was designed to understand spatial context within images and the contrastive loss based pretraining was designed to understand the similarities between images/local regions within images. We observed that self-supervised pretraining mechanisms can reduce the requirements for expensive image annotations in MR imaging applications, thereby reducing the burden on expert image annotations. For the liver segmentation task, we observed that we can match the performance of a fully supervised DL model using only 30% of the labeled data using contrastive loss based pretraining. Using a pretraining mechanism such as contrastive loss also helped improve the performance compared to the supervised baseline when all available labeled training data was used. We also observed that DL model pretrained with contrastive loss mechanism performed better than the DL model pretrained to solve a regression-based task.

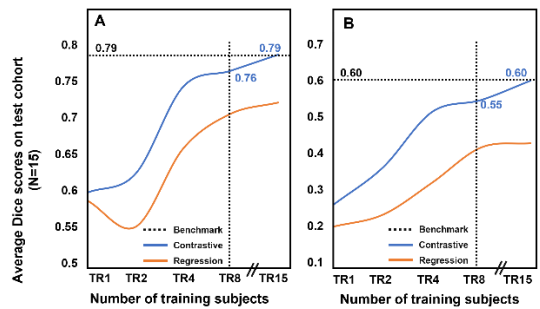

Figure 5: Comparison of segmentation performance of the two self-supervised pretraining techniques on (A) central gland and (B) peripheral zone segmentation task. The Dice scores are averaged over the test cohort (N=15). The dotted line represents the average Dice score from a supervised baseline deep learning model trained on all available labeled training data.

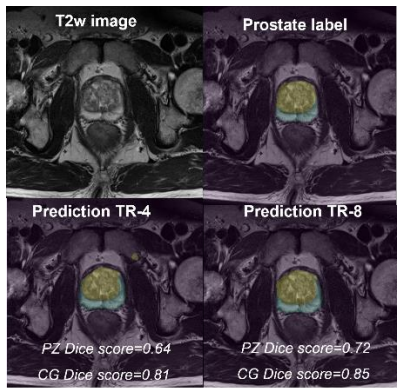

Figure 6: Prostate segmentation. Qualitative comparison of central gland and peripheral zone mask predictions from deep learning model pretrained with contrastive loss and finetuned with four (bottom left) and eight (bottom right) training subjects respectively. The masks for central gland (olive) and peripheral zone (cyan) are overlaid on T2-weighted images of the prostate. The Dice scores for the 3D volume for each case is also displayed.

ACKNOWLEDGMENT

The authors would like to acknowledge grant support from NIH (CA245920), the Arizona Biomedical Research Commission (CTR056039), and the Technology and Research Initiative Fund (TRIF).